\documentclass{article}[submission]
\usepackage[english]{babel}
\usepackage[preprint]{neurips_2020}

\usepackage[utf8]{inputenc} % allow utf-8 input
\usepackage[T1]{fontenc}    % use 8-bit T1 fonts
\usepackage{hyperref}       % hyperlinks
\usepackage{url}            % simple URL typesetting
\usepackage{booktabs}       % professional-quality tables
\usepackage{amsfonts}       % blackboard math symbols
\usepackage{nicefrac}       % compact symbols for 1/2, etc.
\usepackage{microtype}      % microtypography
\usepackage{siunitx} % For scientific notation
\usepackage{chngcntr} % For getting figure counters right
\usepackage[table]{xcolor} % allows alternating table colous
\usepackage{graphicx}
\usepackage{svg}[inkscapelatex=false]
\graphicspath{{./figures/}}    
\usepackage{mathtools}

\title{Conifer Seedling Detection in UAV-Imagery with RGB-Depth Information}

\author{
  Jason Jooste,\textsuperscript{\rm 1}
  Michael Fromm,\textsuperscript{\rm 2} 
  Matthias Schubert\textsuperscript{\rm 2} \\

    \textsuperscript{\rm 1} LMU Munich, Germany \\
    \textsuperscript{\rm 2} Database Systems and Data Mining, LMU Munich, Germany \\
}

\begin{document}

\maketitle

\begin{abstract}
 Monitoring of reforestation is currently being considerably streamlined through the use of drones and image recognition algorithms, which have already proven to be effective on colour imagery. In addition to colour imagery, elevation data is often also available. The primary aim of this work was to improve the performance of the faster-RCNN object detection algorithm by integrating this height information, which showed itself to notably improve performance. Interestingly, the structure of the network played a key role, with direct addition of the height information as a fourth image channel showing no improvement, while integration after the backbone network and before the region proposal network led to marked improvements. This effect persisted with very long training regimes. Increasing the resolution of this height information also showed little effect. 
\end{abstract}

\section{Introduction}

Fossil fuel exploration in the boreal forest of Alberta, Canada has significantly impacted the local environment. Seismic exploration lines are 5-10 m wide and sometimes tens of kilometres long corridors cut through the forest for mining and gas exploration and are one important example of such impacts. These narrow forest clearings have measurably affected the caribou population by allowing access for their predators through the once dense forest \citep{caribou}. This contributed to the Canadian government requiring that companies reforest the affected areas after use \citep{caribouRegulation}. Compliance is measured by, among other factors, the size and number of seedlings growing in the area. The time-consuming process of assessing compliance could be markedly streamlined through the use of drones and subsequent object detection on the recorded imagery.

The detection of individual trees within forests is well estabilished \citep{tree_review_ke_2011}, but the detection of seedlings in particular comes with a number of unique challenges, mostly related to their smaller profile and resulting higher image resolution. \cite{Feduck_canSeedDetection} analysed 3cm GSD colour imagery and achieved a sensitivity of 75\% and specificity of >99\% using image segmentation and classification trees. \cite{frommRemote} surveyed modern object detection algorithms based on convolutional neural networks (CNNs) in the context of seedling detection. They found two-stage networks to perform better than single shot networks, with the Resnet-101 Faster-RCNN model pretrained on the COCO dataset performing best, with a Mean Average Precision (MAP) of 0.81. However, to the best of our knowledge, the addition of height information, which is often either already available for areas under study or increasingly able to be measured by the drones themselves, has not yet been attempted. 

% Hirschmugl et al summary
% Wallace et al study
\subsection{Contributions}
In this work, we integrate height information into traditional object detection by modifying an FPN-Faster-RCNN \citep{FPN} to improve the accuracy of conifer seedling detection. We discovered that accounting for structural properties of the model is important for improving performance and investigated a number of other model and data properties that showed little effect. As height information is often readily available for the areas under study, we consider its integration to be a straightforward approach to improving conifer seedling detection. Accurate detection of conifer seedlings could allow for a more streamlined monitoring of reforestation efforts. 
\section{Related Work}
The following section revisits related work within the field of seedling detection and general object detection with additional height information on overhead imagery.

\subsection{Seedling Detection}
The work from \cite{rs11212585} first applied convolutional neural networks (CNNs) for seedling detection on images shoot by Unmanned Aerial Vehicles (UAVs). They used multiple object detectors (e.g. Faster-RCNN \cite{fasterRCNN}) on the backbone of a Res-Net 101 \cite{resnet} and achieved a performance of about 81 \% mean average precision. \cite{SCHIEFER2020205} deviated from the original task by additional classes (introduction of nine tree species, three genus-level classes, deadwood and the forest floor) and achieved an mean F1-score of 0.73. They used an semantic segmentation approach (U-net) \cite{li2017multi}. \cite{PEARSE2020156} studied two trial sites containing over 30,000 seedlings with CNN-based models. With an model trained on seedlings from both sites, they achieved 99.4 \% sensitivity and a precision of 97\%.

\begin{figure}
    \centering
    \includegraphics[width=0.8\textwidth]{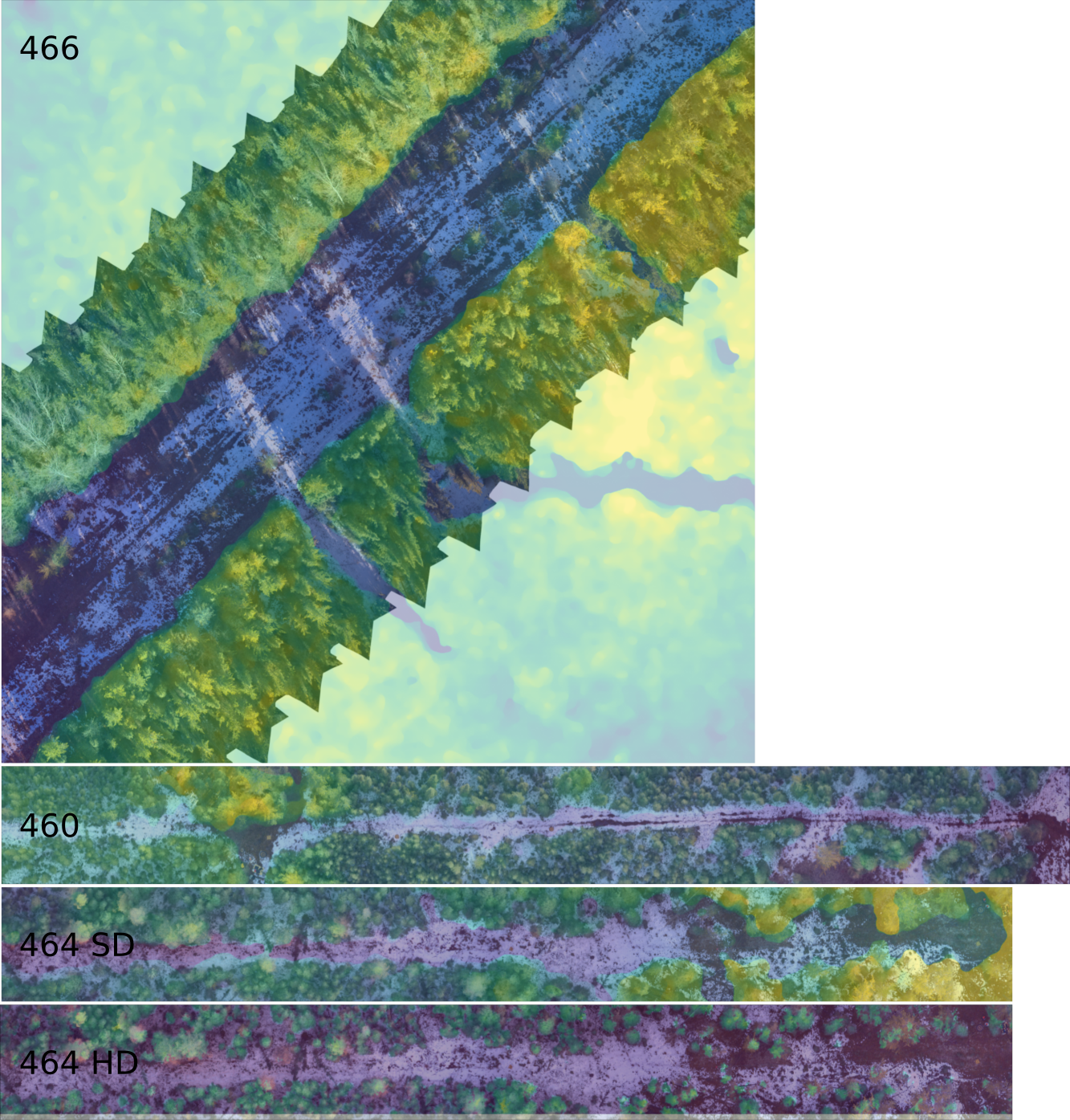}
    \caption{The lanes used in this data analysis. All lanes show the RGB recorded image data overlayed with height data from low (purple) to high (yellow). Lane 466 clearly shows the boundaries of the recorded image data, with only height data available beyond the boundaries.}
    \label{fig:lanes}
\end{figure}

\section{Dataset}
The study area analysed in this work was the boreal forest of north east Alberta, Canada. The dataset consists of images taken from a DJI Mavic Pro Drone over \textasciitilde 25m sections of three seismic lines (or "Lanes") hereby referred to as Lane 460, Lane 464 and Lane 466. All images were taken in October 2017 (leaf off) at a height of \textasciitilde 30m with a Ground Sampling Distance (GSD) of \textasciitilde 0.75cm. The individual drone images were rectified into a single image for each lane and then split into 256x256 pixel image tiles, which served as the data for testing and training of the network. In the uncommon case that a seedling was split across multiple images, it was considered to be two separate seedlings. As a result of the orientation of Lane 466, images outside of the recorded area were removed by filtering out tiles containing more than 5\% pure white. A composite of the height and image data can be seen in Figure \ref{fig:lanes}.

Height data for the lanes was extracted from a photogrammetric Digital Surface Model (DSM) of the larger research area taken by a fixed wing plane also in October 2017. Higher resolution data was additionally available for Lane 464 in the form of a Canopy Height Model (CHM) that included information from a LiDAR-based digital terrain model. The generation of ground truth bounding boxes was performed manually with the software LabelImg \cite{LabelImg}, with early annotations serving as an initial training to precompute suggestions for further manual annotation. Many images, referred to as negative images or negative samples, did not contain any seedlings at all (see Table \ref{tab:data}) and needed to be addressed separately in the cross-validation process. 

\begin{table}
    \centering
    \caption{A summary of the 256x256 images and ground truth seedling detections in each lane.}
    {\rowcolors{3}{white}{gray!10}
    \begin{tabular}{lrrrr}
\toprule
{} & \multicolumn{3}{l}{Images} &   Annotations \\
{} &    +ve &   -ve & Total & Num seedlings \\
Lane  &        &       &       &               \\
\midrule
460   &    135 &   643 &   778 &           277 \\
464   &    186 &   552 &   738 &           537 \\
466   &    199 &  2150 &  2349 &           436 \\
Total &    520 &  3345 &  3865 &          1250 \\
\bottomrule
\end{tabular}

    }
    \label{tab:data}
\end{table}
\section{Model architectures}

The baseline (Vanilla) model was derived from the FasterRCNN model with Resnet50-FPN backbone from the TorchVision \citep{torchvision} implemented in Pytorch \citep{pytorch}. The Resnet50-FPN model is based on an adapted version of the the Faster-RCNN object detection model presented in \cite{fasterRCNN}, where \cite{FPN} improved the standard Faster-RCNN by adding a Feature Pyramid Network (FPN) on top of the Resnet-50 \citep{resnet} backbone. The Resnet50-FPN model consists of four main components: the Backbone, the Region Proposal Network (RPN), Region of Interest pooling (RoI-Pooling) and the Region of Interest Heads (RoI-Heads). 

The Backbone component is designed to take the original image and embed it in a more meaningful feature space using a pretrained network. The three-channel input image through an Resnet-50-FPN that returns five feature maps of varying resolution. This can be seen in more detail in Appendix \ref{fig:FPN} or the original work \citet{FPN}. This is followed by the Region Proposal Network (RPN), a CNN designed to detect regions of the image that likely contain relevant information for the subsequent object classification task. For each pixel of the input feature map, it determines the probability that a set of pre-defined bounding boxes centred upon the pixel contain an object. A balanced set of positive and negative proposals are then passed individually through RoI pooling. This is followed by the RoI heads, which consists of two dense layers and finally a classification and "bounding-box regression" heads to generate the model outputs. 

Transfer learning is commonly applied to this process, in that the weights of the backbone are taken from a model pretrained on the ImageNet classification task. This is based on the assumption that the feature representation learnt on the more data-rich image recognition task is also useful in the object detection setting.  As an extension to using a pretrained backbone, it is also possible to take an entire object detection model pretrained on the COCO object detection dataset. In this case, the only modifications required are to replace the final classification layer with a dense layer with two outputs, as the COCO dataset contains 91 different classes and our task involves only binary classification. In the models without pretraining, all weights were initialised randomly, otherwise a FasterRCNN network pretrained on the COCO dataset was used. 

\begin{figure}
    \centering
    \includegraphics[width=\columnwidth]{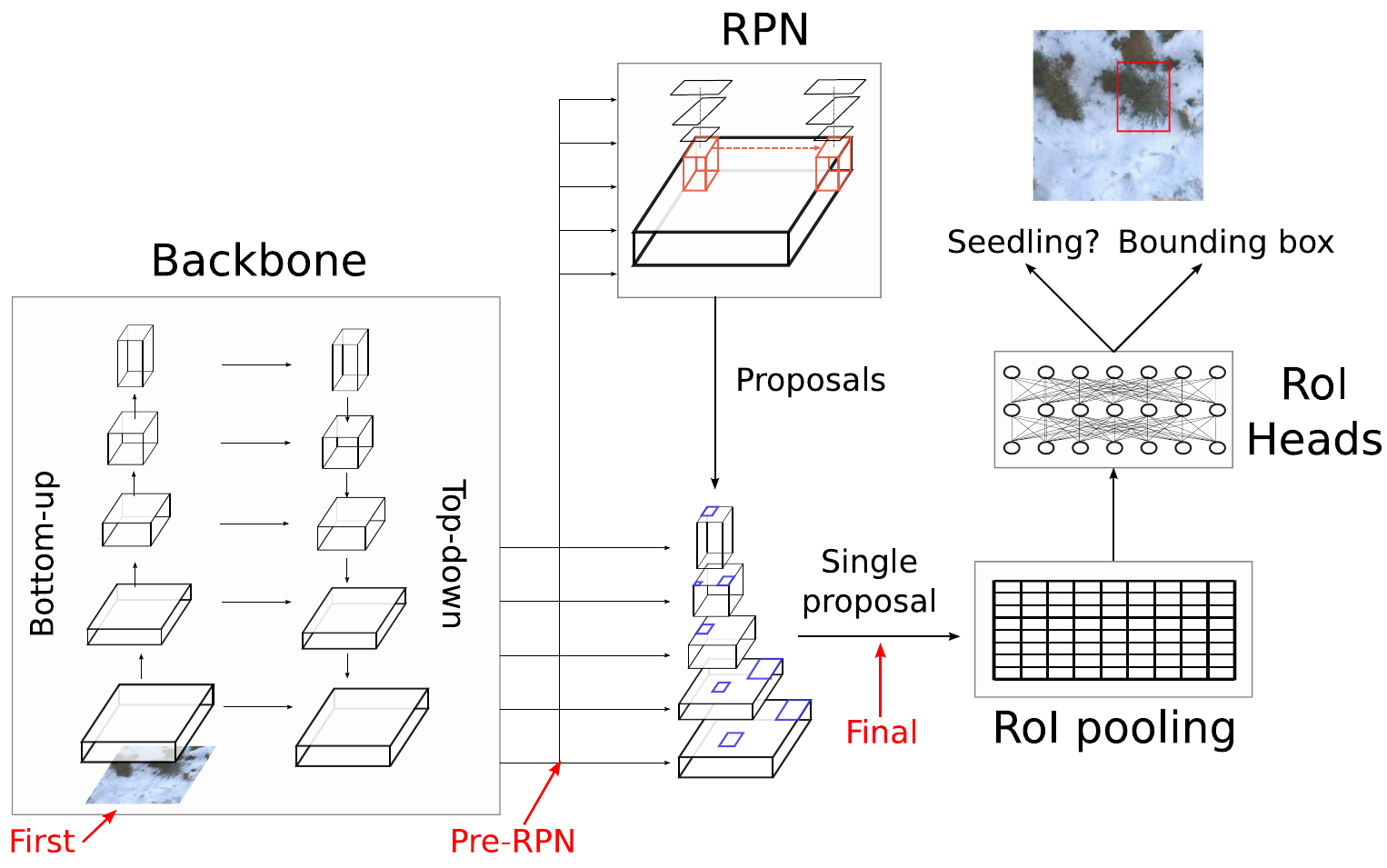}
    \caption{A diagram of the flow of information through the Vanilla baseline network. The image (bottom left) passes through the bottom-up and top-down pathways of the FPN backbone, generating five feature maps. These feature maps are all used by the RPN to generate five sets of proposals, which are then scaled. These proposals define the areas of the feature map that are taken into RoI pooling, which pools the five feature maps into a consistent 7x7x256 feature map, which is then passed to the RoI Heads. The RoI heads consists of two dense layers, that then splits into two further sets of two dense layers - one for the classification of the region as either seedling or background and one for bounding box regression. Red text shows where modifications for each new model class were made.}
    \label{fig:model_diag}
\end{figure}

\subsection{Model modifications}

The "First" model integrates height information directly as a fourth channel to the original image. The only modification required was to add an extra row to the first convolutional layer of the backbone. The "Final" model was generated by performing RoI pooling on the full-scale height data for each proposal and then concatenating the pooled height information to the existing pooled feature map. The weights of the first dense layer of the RoI heads were extended to account for this new information. To create the "Pre-RPN" model, the height information was concatenated directly to the inputs of both the RPN and RoI Heads. To accommodate this change, the weights in both the RPN and RoI Heads were extended.

Variants of the Pre-RPN models were investigated with different backbone feature map depths. The number is controlled by the FPN structure itself (and doesn't require retraining of the backbone) and the depth is controlled by the final convolutional layer. The first convolutional layer of the RPN and the first dense layer of the RoI-Heads were adapted accordingly as above. In all models, any new weights added to the network were initialised with Glorot initialisation \citep{Glorot2010}. 
\section{Experiment}

The performance of the models was evaluated with repeated random subset crossvalidation, with the dataset randomly split into five different 80/20 train and test sets. A key factor in the analysis was the ratio of positive to negative samples, that is, the ratio of images that contained at least one seedling (positive samples) to those that did not contain any seedlings (negative samples). In the dataset, the number of negative samples (3345) vastly outweighed the number of positive samples (520). For experiments 1, 2 and 4 the negative samples were excluded altogether, while in Experiment 3 a ratio of 0.5 negative to positive samples was enforced and for Experiment 5 all images were used for training and testing. For each CV split, each model was run multiple times with a range of random seeds, to give an indication of the variance in the cross validation score, which was averaged across the five cross validation splits. A further 10\% of the train set was set aside during runtime as a validation set, which was used to evaluate model performance after each epoch. The best performing model was selected as representative for a given hyperparameter configuration.

Model performance was evaluated with the Mean Average Precision (MAP) at an IoU of 0.5, a common performance measure in object detection, calculated with the Pycocotools library. The library approximates the MaP by calculating the precision values for 101 different recall scores and is detailed more clearly in \citep{cocoGit}. It is worth noting that, as there are only two classes in this case (Seedling and background), the MAP is equivalent to the average precision. All models were all trained by Stochastic Gradient Descent (SGD) with a batch size of 5, a learning rate of 2e-3, a momentum of 0.9 and a weight decay of 1e-4, which were determined through limited experimentation to be reasonable parameters for these models and data. No hyperparameter tuning was performed and this could certainly lead to improved performance. As we encountered occasional instances of exploding gradients, the L2 norm was clipped at a value of 1.

Five key experiments were performed. The first experiment concerned the performance of the different model types and the models (Vanilla, First, Final and Pre-RPN) were all tested with and without pretraining, trained for 35 and 100 epochs respectively. Each model was tested with five different seeds per CV split to give a total of 200 trained models. 

Experiment 2 involved a comparison between different backbone output feature depths of the Pre-RPN model. Six models of depth 16, 32, 64, 128, 192, 256 were tested with cross validation and five different seeds giving a total of 450 trained models. 

The third experiment concerned the effect of height data resolution on the performance of the Pre-RPN model. The pretrained and non-pretrained Pre-RPN models trained and tested on either high resolution data (HD) or lower resolution (SD) height data of Lane 464. These models were also trained on five CV splits for each resolution for five different seeds giving a total of 100 trained models. 

The fourth experiment involved testing the First, Vanilla and Pre-RPN models under longer training regimes. Training was allowed to continue for 150 epochs for all models with and without pretraining. Only two cross validation splits and two seeds were applied to reduce training time giving a total of 24 trained models. 

The fifth experiment was designed to investigate the effect of negative samples on the performance of the Pre-RPN model. The pretrained Pre-RPN model and Vanilla models were trained on five CV splits containing all negative samples for five seed values giving a total of 50 trained models. 
\section{Results}
\begin{table}
    \centering
    \caption{The mean test MAP of all models in all experiments. FM Layers describes the number of layers outputted by the FPN backbone. -/+ describes the number of negative samples per positive sample in the dataset, with "all" indicating that all images were used. SD describes the standard deviation in the cross validated mean test MAP across the different seeds, i.e. after all cross validation splits with the same seed were averaged.}
    % \rowcolors{5}{white}{gray!25}
\begin{tabular}{llllllllrr}
\toprule
Exp & Model & Pretrain & \begin{tabular}{l} FM  \\ Layers \end{tabular} & Lane & Res. & -/+ & Epochs & \begin{tabular}{l} Test \\ MAP \end{tabular} &     SD     \\
\midrule
\rowcolor{gray!10} Exp 1 & Final & False & 256 & 460/4/6 & SD & 0 & 100 &     0.286 &    0.013  \\
\rowcolor{gray!10}      & First & False & 256 & 460/4/6 & SD & 0 & 100 &     0.266 &    0.015 \\
\rowcolor{gray!10}      & Pre-RPN & False & 256 & 460/4/6 & SD & 0 & 100 &     \textbf{0.612} &    0.013 \\ 
\rowcolor{gray!10}      & Vanilla & False & 256 & 460/4/6 & SD & 0 & 100 &     0.289 &    0.009 \\
       & Final & True  & 256 & 460/4/6 & SD & 0 & 35  &     0.570 &    0.009 \\
      & First & True  & 256 & 460/4/6 & SD & 0 & 35  &     0.571 &    0.011 \\
      & Pre-RPN & True  & 256 & 460/4/6 & SD & 0 & 35  &     \textbf{0.704} &    0.009 \\
      & Vanilla & True  & 256 & 460/4/6 & SD & 0 & 35  &     0.579 &    0.014 \\
\rowcolor{gray!10} Exp 2 & Pre-RPN & True  & 16  & 460/4/6 & SD & 0 & 35  &     0.708 &    0.004 \\
\rowcolor{gray!10}      &         &       & 32  & 460/4/6 & SD & 0 & 35  &     \textbf{0.721} &    0.006 \\
\rowcolor{gray!10}      &         &       & 64  & 460/4/6 & SD & 0 & 35  &     0.709 &    0.008 \\
\rowcolor{gray!10}      &         &       & 128 & 460/4/6 & SD & 0 & 35  &     0.715 &    0.009 \\
\rowcolor{gray!10}      &         &       & 192 & 460/4/6 & SD & 0 & 35  &     0.714 &    0.012 \\
\rowcolor{gray!10}      &         &       & 256 & 460/4/6 & SD & 0 & 35  &     0.704 &    0.007 \\
 Exp 3 & Pre-RPN & False & 256 & 464 & HD & 0.5 & 35  &     \textbf{0.563} &    0.007 \\
      &         &       &     &         & SD & 0.5 & 35  &     0.544 &    0.014 \\
\rowcolor{gray!10}       &         & True  & 256 & 464 & HD & 0.5 & 35  &     \textbf{0.605} &    0.011 \\
 \rowcolor{gray!10}     &         &       &     &         & SD & 0.5 & 35  &     0.601 &    0.016 \\
Exp 4 & First & False & 256 & 460/4/6 & SD & 0 & 150 &     0.324 &    0.014 \\
      & Pre-RPN & False & 256 & 460/4/6 & SD & 0 & 150 &     \textbf{0.652} &    0.000 \\
      & Vanilla & False & 256 & 460/4/6 & SD & 0 & 150 &     0.324 &    0.018 \\
 \rowcolor{gray!10}      & First & True  & 256 & 460/4/6 & SD & 0 & 150 &     0.593 &    0.016 \\
\rowcolor{gray!10}      & Pre-RPN & True  & 256 & 460/4/6 & SD & 0 & 150 &     \textbf{0.718} &    0.026 \\
\rowcolor{gray!10}      & Vanilla & True  & 256 & 460/4/6 & SD & 0 & 150 &     0.605 &    0.001 \\
Exp 5 & Pre-RPN & True  & 256 & 460/4/6 & SD & all & 35  &     \textbf{0.537} &    0.012 \\
      & Vanilla & True  & 256 & 460/4/6 & SD & all & 35  &     0.475 &    0.013 \\
\bottomrule
\end{tabular}

    \label{tab:big_table}
\end{table}
% Manual table alterations: Add bold. Add \rowcolor{gray!10} to the different parts of the experiment to be compared. Column names should be updated in export notebook.

Table \ref{tab:big_table} shows that the Pre-RPN models performed considerably better than all other models in Experiment 1 with and without pretraining. Less pretraining strengthened this effect, with the Pre-RPN model showing approximately a mean improvement of greater than 0.3 in the MAP with respect to all other models without pretraining. The First and Final models did not improve performance, even appearing to cause slight decreases. We speculate that the poor performance of the First model could be explained by the loss of this new information through the very large and deep backbone, which has already been pretrained for a different task. The poor performance of the Final model could be explained by the lack of context. That is, in the final layer, the height information is fed directly into a fully-connected layer, without passing first through a CNN. This does not allow the integration of local interactions between the visual features and local height into the final feature representation. 
 
Experiment 2 showed that backbone output channel depth may play a role in model performance, with the standard depth of 256 showing the worst performance. There was however no clear trend with the models with 16 and 64 layers also showing the poor performance and the 32 layer model showing the best performance. This increase may simply be a consequence of the lower relevance of the new height channel with increasing numbers of layers and could be influenced by how the new weights introduced to process this channel were initialised, with higher initial weights possibly counteracting this effect. This effect could be investigated with further hyperparameter tuning.

In Experiment 3, an increase in height data resolution showed a small improvement in performance of the models without pretraining and no clear benefit for models with pretraining. This suggests that low-resolution information about the location of the image (in lane or out of lane, for example) may be enough to improve performance and any further detail may prove to be unnecessary. Further investigations with more data would be informative in this respect. Experiment 4 further showed that even with longer training times, the first model was not able to integrate the height information and improve model performance over the baseline.

Finally, Experiment 5 investigated the performance of the pretrained Pre-RPN and Vanilla models on the entire dataset, with a large bias towards negative samples. This shows that the Pre-RPN models have an average improvement upon the Vanilla models in MAP of approximately 0.06, which is less approximately half of the improvement with the only-positive dataset. The reduction in performance is a direct result of the model evaluation criterion, with the MAP being a monotonically non-increasing function with increasing negative samples. This could also be a cause of the reduced improvement of the new model class. 

\subsection{Future work}
This work was intended to establish the possibility of improving performance with height information and no significant effort was therefore made to maximise ultimate model performance. However, there are a large number of technical improvements that could be made to the model should absolute model performance be of importance. One crucial aspect is data augmentation. In small datasets such as these, augmenting the existing data by, for example, flipping, cropping, rotation or colour adjustment, can increase effective training set size and boost performance, as in \cite{frommRemote}. Further hyperparameter tuning of network structure (as evidenced by the improved performance with FPN feature map depth of 32) could lead to further improvements. An image size of 256x256 also showed itself to be suboptimal when combined with the Faster-RCNN. As the Faster-RCNN was pretrained on the COCO dataset, it requires a minimum image size of 800x800, with any smaller images scaled up to this size. This scaling could be avoided, and computation time could be improved, by natively using image segments of this size. Furthermore, considerable improvements in training speed of the Pre-RPN model might be possible if the layers of the Restnet backbone (the bulk of the parameters in the model) were frozen, as the newly introduced height information does not interact with them. As mentioned in the Pre-RPN model analysis, experimentation with the weight initialisation for the height information could also improve performance and training speed. Repeated random subset cross validation was also suboptimal as there are some samples that never appeared in the test set of any CV split, while others appeared multiple times. Simple k-fold cross validation would likely less variation in MAP as all datapoints are forced to appear exactly once in the test sets. 

On a broader scale, a key opportunity may lie in the combination of various network structures to make the application of this detection method more general. Models are available for the prediction of the height of a pixel based only on rgb images. If this approach were to be combined with the models in this work, there could possibly lead to performance improvement, even if true height data is not available. Superresolution techniques have already been applied to the detection of conifer seedlings \cite{superresolution} and which may allow for lower resolution imagery to be upscaled and then supplied to this algorithm, allowing the data-collecting drones to fly at higher altitudes and cover more ground in less time - a key factor in the economic feasability of these monitoring methods. This is of particular interest as the results of Experiment 2 suggest that height data quality may have little impact on performance improvements.

\section{Conclusion}
This work shows that considerable improvements in conifer seedling detection are possible by including height information into the model. The location of height information input proved to play an important role in model performance. Of the three approaches: First, as a fourth channel to the image; Final, added to the final fully connected layer and Pre-RPN, before the region proposal network and RoI pooling, only the Pre-RPN model showed an increased performance, both with and without pretraining. The quality of the height information appears to play little role, however, in the performance of the model. Multiple FPN backbone output depths were tested and a depth of 256 appears to be suitable, though there is a suggestion that lower channel depth may lead to better performance. The data suggests that the integration of height information into seedling detection models can lead to an appreciable improvement in performance, which could aid the monitoring of reforestation efforts in Canada and elsewhere. 

\section{Acknowledgements}
This research is part of the Boreal Ecosystem Recovery and Assessment (BERA) project (\url{www.bera-project.org}), and was supported by a Natural Sciences and Engineering Research Council of Canada Collaborative Research and Development Grant (CRDPJ 469943-14) in conjunction with Alberta-Pacific Forest Industries, Cenovus Energy, ConocoPhillips Canada and Canadian Natural Resources Ltd.

\clearpage
\bibliographystyle{abbrvnat}
\bibliography{references}

\begin{thebibliography}{18}
\providecommand{\natexlab}[1]{#1}
\providecommand{\url}[1]{\texttt{#1}}
\expandafter\ifx\csname urlstyle\endcsname\relax
  \providecommand{\doi}[1]{doi: #1}\else
  \providecommand{\doi}{doi: \begingroup \urlstyle{rm}\Url}\fi

\bibitem[Beguin et~al.(2013)Beguin, McIntire, Fortin, Cumming, Raulier, Racine,
  and Dussault]{caribou}
J.~Beguin, E.~J.~B. McIntire, D.~Fortin, S.~G. Cumming, F.~Raulier, P.~Racine,
  and C.~Dussault.
\newblock Explaining geographic gradients in winter selection of landscapes by
  boreal caribou with implications under global changes in eastern canada.
\newblock \emph{PLOS ONE}, 8\penalty0 (10):\penalty0 null, 10 2013.
\newblock \doi{10.1371/journal.pone.0078510}.
\newblock URL \url{https://doi.org/10.1371/journal.pone.0078510}.

\bibitem[Feduck et~al.(2018)Feduck, McDermid, and
  Castilla]{Feduck_canSeedDetection}
C.~Feduck, G.~J. McDermid, and G.~Castilla.
\newblock Detection of coniferous seedlings in uav imagery.
\newblock \emph{Forests}, 9\penalty0 (7), 2018.
\newblock ISSN 1999-4907.
\newblock \doi{10.3390/f9070432}.
\newblock URL \url{https://www.mdpi.com/1999-4907/9/7/432}.

\bibitem[Fromm et~al.(2019{\natexlab{a}})Fromm, Berrendorf, Faerman, Chen,
  Schüss, and Schubert]{superresolution}
M.~Fromm, M.~Berrendorf, E.~Faerman, Y.~Chen, B.~Schüss, and M.~Schubert.
\newblock Xd-stod: Cross-domain superresolution for tiny object detection.
\newblock In \emph{2019 International Conference on Data Mining Workshops
  (ICDMW)}, pages 142--148, 2019{\natexlab{a}}.
\newblock \doi{10.1109/ICDMW.2019.00031}.

\bibitem[Fromm et~al.(2019{\natexlab{b}})Fromm, Schubert, Castilla, Linke, and
  McDermid]{frommRemote}
M.~Fromm, M.~Schubert, G.~Castilla, J.~Linke, and G.~McDermid.
\newblock Automated detection of conifer seedlings in drone imagery using
  convolutional neural networks.
\newblock \emph{Remote Sensing}, 11\penalty0 (21), 2019{\natexlab{b}}.
\newblock ISSN 2072-4292.
\newblock \doi{10.3390/rs11212585}.
\newblock URL \url{https://www.mdpi.com/2072-4292/11/21/2585}.

\bibitem[Fromm et~al.(2019{\natexlab{c}})Fromm, Schubert, Castilla, Linke, and
  McDermid]{rs11212585}
M.~Fromm, M.~Schubert, G.~Castilla, J.~Linke, and G.~McDermid.
\newblock Automated detection of conifer seedlings in drone imagery using
  convolutional neural networks.
\newblock \emph{Remote Sensing}, 11\penalty0 (21), 2019{\natexlab{c}}.
\newblock ISSN 2072-4292.
\newblock \doi{10.3390/rs11212585}.
\newblock URL \url{https://www.mdpi.com/2072-4292/11/21/2585}.

\bibitem[Glorot and Bengio(2010)]{Glorot2010}
X.~Glorot and Y.~Bengio.
\newblock Understanding the difficulty of training deep feedforward neural
  networks.
\newblock In Y.~W. Teh and D.~M. Titterington, editors, \emph{AISTATS},
  volume~9 of \emph{JMLR Proceedings}, pages 249--256. JMLR.org, 2010.
\newblock URL
  \url{http://dblp.uni-trier.de/db/journals/jmlr/jmlrp9.html#GlorotB10}.

\bibitem[He et~al.(2016)He, Zhang, Ren, and Sun]{resnet}
K.~He, X.~Zhang, S.~Ren, and J.~Sun.
\newblock Deep residual learning for image recognition.
\newblock In \emph{2016 IEEE Conference on Computer Vision and Pattern
  Recognition (CVPR)}, pages 770--778, 2016.
\newblock \doi{10.1109/CVPR.2016.90}.

\bibitem[Ke and Quackenbush(2011)]{tree_review_ke_2011}
Y.~Ke and L.~J. Quackenbush.
\newblock A review of methods for automatic individual tree-crown detection and
  delineation from passive remote sensing.
\newblock \emph{International Journal of Remote Sensing}, 32\penalty0
  (17):\penalty0 4725--4747, 2011.
\newblock \doi{10.1080/01431161.2010.494184}.
\newblock URL \url{https://doi.org/10.1080/01431161.2010.494184}.

\bibitem[Li et~al.(2017)Li, Sarma, Ho, Gertych, Knudsen, and
  Arnold]{li2017multi}
J.~Li, K.~V. Sarma, K.~C. Ho, A.~Gertych, B.~S. Knudsen, and C.~W. Arnold.
\newblock A multi-scale u-net for semantic segmentation of histological images
  from radical prostatectomies.
\newblock In \emph{AMIA Annual Symposium Proceedings}, volume 2017, page 1140.
  American Medical Informatics Association, 2017.

\bibitem[Lin(2021)]{LabelImg}
T.~T. Lin.
\newblock Labelimg.
\newblock \url{https://github.com/tzutalin/labelImg}, 2021.

\bibitem[Lin et~al.(2015)Lin, Maire, Belongie, Hays, Perona, Ramanan,
  Doll{\'a}r, Zitnick, Fleet, Pajdla, Schiele, and Tuytelaars]{cocoGit}
T.-Y. Lin, M.~Maire, S.~Belongie, J.~Hays, P.~Perona, D.~Ramanan,
  P.~Doll{\'a}r, C.~L. Zitnick, D.~Fleet, T.~Pajdla, B.~Schiele, and
  T.~Tuytelaars.
\newblock cocoapi.
\newblock \url{https://github.com/cocodataset/cocoapi}, 2015.

\bibitem[Lin et~al.(2017)Lin, Dollár, Girshick, He, Hariharan, and
  Belongie]{FPN}
T.-Y. Lin, P.~Dollár, R.~Girshick, K.~He, B.~Hariharan, and S.~Belongie.
\newblock Feature pyramid networks for object detection.
\newblock In \emph{2017 IEEE Conference on Computer Vision and Pattern
  Recognition (CVPR)}, pages 936--944, 2017.
\newblock \doi{10.1109/CVPR.2017.106}.

\bibitem[Marcel and Rodriguez(2010)]{torchvision}
S.~Marcel and Y.~Rodriguez.
\newblock Torchvision the machine-vision package of torch.
\newblock In \emph{Proceedings of the 18th ACM International Conference on
  Multimedia}, MM '10, page 1485–1488, New York, NY, USA, 2010. Association
  for Computing Machinery.
\newblock ISBN 9781605589336.
\newblock \doi{10.1145/1873951.1874254}.
\newblock URL \url{https://doi.org/10.1145/1873951.1874254}.

\bibitem[Paszke et~al.(2019)Paszke, Gross, Massa, Lerer, Bradbury, Chanan,
  Killeen, Lin, Gimelshein, Antiga, Desmaison, Kopf, Yang, DeVito, Raison,
  Tejani, Chilamkurthy, Steiner, Fang, Bai, and Chintala]{pytorch}
A.~Paszke, S.~Gross, F.~Massa, A.~Lerer, J.~Bradbury, G.~Chanan, T.~Killeen,
  Z.~Lin, N.~Gimelshein, L.~Antiga, A.~Desmaison, A.~Kopf, E.~Yang, Z.~DeVito,
  M.~Raison, A.~Tejani, S.~Chilamkurthy, B.~Steiner, L.~Fang, J.~Bai, and
  S.~Chintala.
\newblock Pytorch: An imperative style, high-performance deep learning library.
\newblock In \emph{Advances in Neural Information Processing Systems 32}, pages
  8024--8035. Curran Associates, Inc., 2019.
\newblock URL
  \url{http://papers.neurips.cc/paper/9015-pytorch-an-imperative-style-high-performance-deep-learning-library.pdf}.

\bibitem[Pearse et~al.(2020)Pearse, Tan, Watt, Franz, and Dash]{PEARSE2020156}
G.~D. Pearse, A.~Y. Tan, M.~S. Watt, M.~O. Franz, and J.~P. Dash.
\newblock Detecting and mapping tree seedlings in uav imagery using
  convolutional neural networks and field-verified data.
\newblock \emph{ISPRS Journal of Photogrammetry and Remote Sensing},
  168:\penalty0 156--169, 2020.
\newblock ISSN 0924-2716.
\newblock \doi{https://doi.org/10.1016/j.isprsjprs.2020.08.005}.
\newblock URL
  \url{https://www.sciencedirect.com/science/article/pii/S0924271620302136}.

\bibitem[Preiffer(2016)]{caribouRegulation}
Z.~Preiffer.
\newblock Cumulative effects, caribou and national energy board regulation.
\newblock In \emph{Proceeding presented at the 36th annual meeting of the
  International Association for Impact Assessment, Achi-Nagoya}, 2016.

\bibitem[Ren et~al.(2015)Ren, He, Girshick, and Sun]{fasterRCNN}
S.~Ren, K.~He, R.~Girshick, and J.~Sun.
\newblock Faster r-cnn: Towards real-time object detection with region proposal
  networks.
\newblock In C.~Cortes, N.~Lawrence, D.~Lee, M.~Sugiyama, and R.~Garnett,
  editors, \emph{Advances in Neural Information Processing Systems}, volume~28.
  Curran Associates, Inc., 2015.
\newblock URL
  \url{https://proceedings.neurips.cc/paper/2015/file/14bfa6bb14875e45bba028a21ed38046-Paper.pdf}.

\bibitem[Schiefer et~al.(2020)Schiefer, Kattenborn, Frick, Frey, Schall, Koch,
  and Schmidtlein]{SCHIEFER2020205}
F.~Schiefer, T.~Kattenborn, A.~Frick, J.~Frey, P.~Schall, B.~Koch, and
  S.~Schmidtlein.
\newblock Mapping forest tree species in high resolution uav-based rgb-imagery
  by means of convolutional neural networks.
\newblock \emph{ISPRS Journal of Photogrammetry and Remote Sensing},
  170:\penalty0 205--215, 2020.
\newblock ISSN 0924-2716.
\newblock \doi{https://doi.org/10.1016/j.isprsjprs.2020.10.015}.
\newblock URL
  \url{https://www.sciencedirect.com/science/article/pii/S0924271620302938}.

\end{thebibliography}

\clearpage
\appendix
\counterwithin{figure}{section}
\section{Appendix}
\begin{figure}[h]
    \centering
    \includegraphics[height=0.8\textheight
    ]{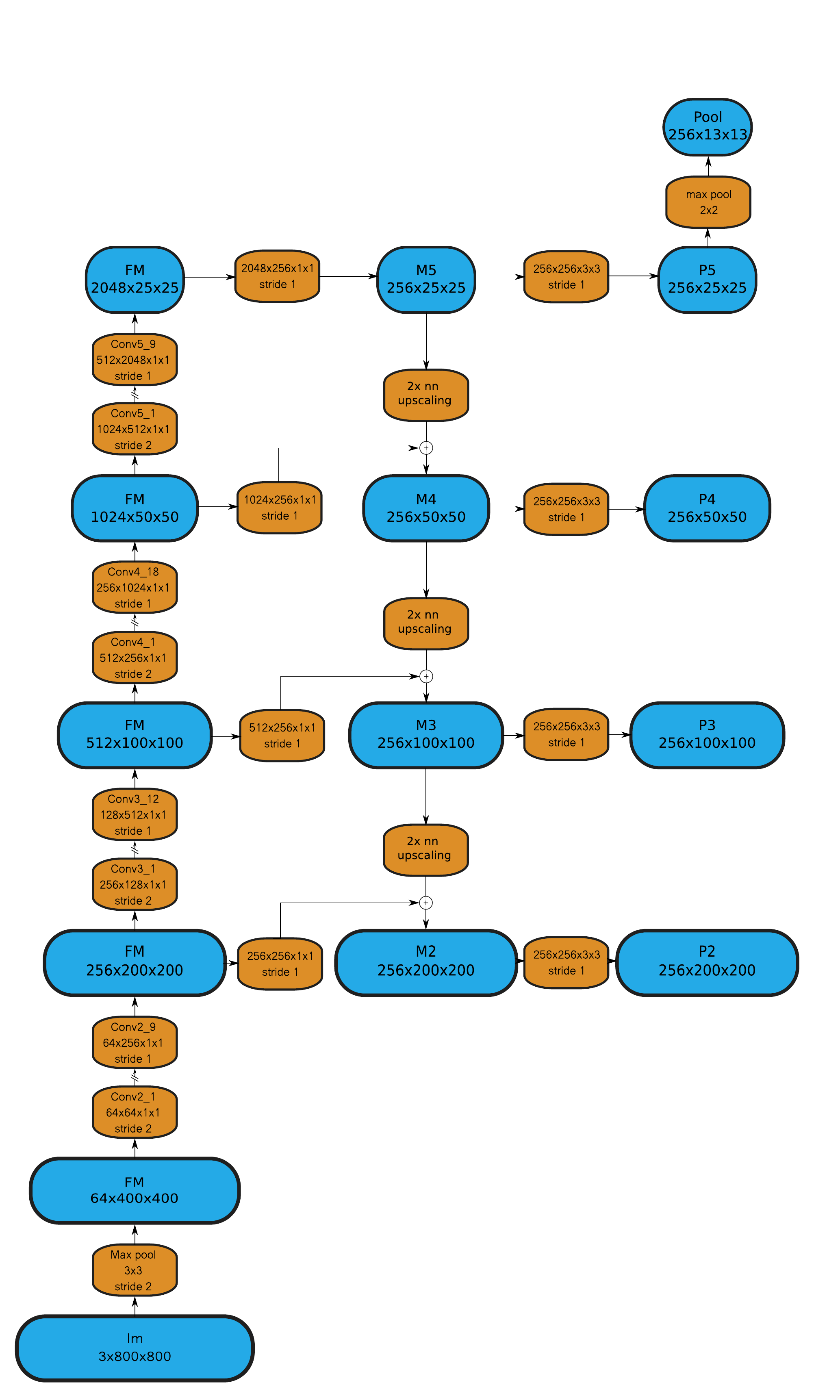}
    \caption{A diagram showing in detail the relevant layers of the Resnet50 FPN used as a backbone for this work}
    \label{fig:FPN}
\end{figure}

\begin{figure}
    \centering
    \includegraphics[width=1.0\textwidth]{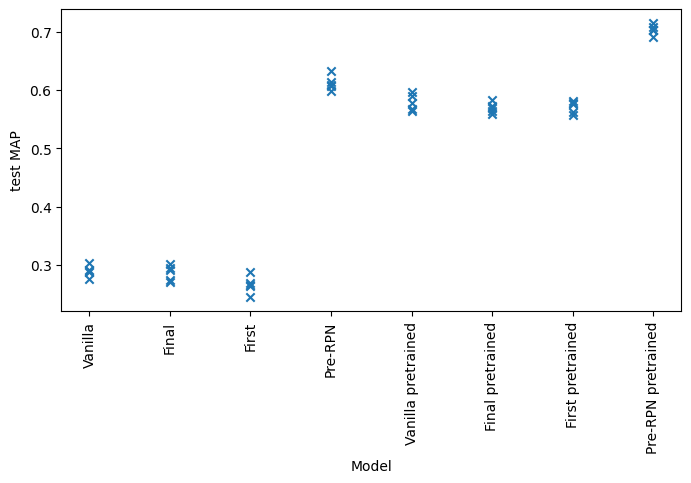}
    \caption{The MAP performance of the models under the three pretraining regimes}
    \label{fig:model_MAPs}
\end{figure}

\begin{figure}
    \centering
    \includegraphics[width=1.0\textwidth]{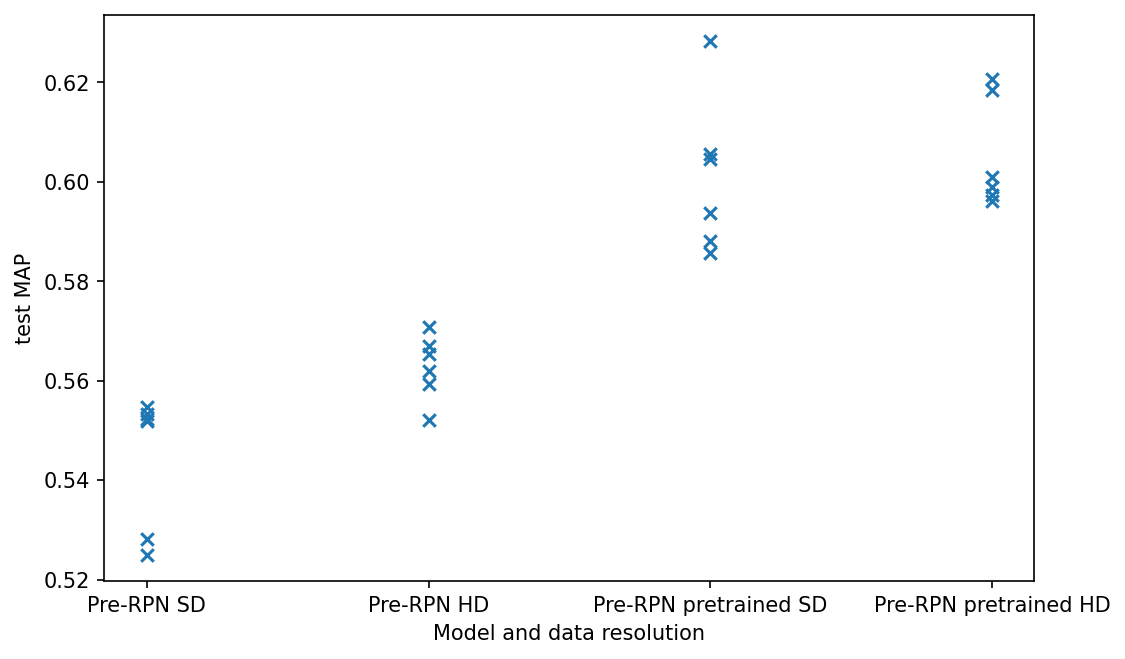}
    \caption{The performance of the Pre-RPN model with high definition (HD) and standard definition (SD) height data under the Pretraining and No-Pretraining schemes}
    \label{fig:HD_SD}
\end{figure}

% Discuss the long training bit here

\begin{figure}
    \centering
    \includegraphics[width=1.0\textwidth]{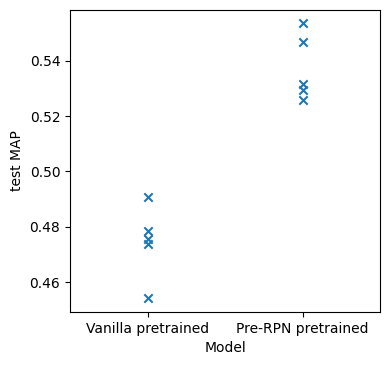}
    \caption{The performance of the pretrained Pre-RPN and Vanilla models on the dataset without any included negative samples}
    \label{fig:no_neg}
\end{figure}
\end{document}